\documentclass{article} %
\usepackage[final]{colm2025_conference}

\usepackage{microtype}
\usepackage{hyperref}
\usepackage{url}
\usepackage{booktabs}

\usepackage{lineno}

\definecolor{darkblue}{rgb}{0, 0, 0.5}
\hypersetup{colorlinks=true, citecolor=darkblue, linkcolor=darkblue, urlcolor=darkblue}

\title{Maximizing Prefix-Confidence at Test-Time \\ Efficiently Improves Mathematical Reasoning}

\author{Matthias Otth, Jonas Hübotter, Ido Hakimi, Andreas Krause \\
ETH Z\"urich, Switzerland
}

\usepackage{amsmath}
\usepackage{amssymb}
\usepackage{mathtools}
\usepackage{amsthm}
\usepackage{bbm}
\usepackage{xspace}
\usepackage{nicefrac}
\usepackage{algorithm}
\usepackage{algpseudocode}  %
\usepackage{wrapfig}
\usepackage{subcaption}
\usepackage{titletoc}
\usepackage{multirow}
\usepackage{xcolor,colortbl}
\usepackage{arydshln}
\usepackage{enumitem}
\definecolor{lightgray}{gray}{0.95}

\usepackage[capitalize,noabbrev]{cleveref}

\usepackage{booktabs}
\usepackage{multirow}

\newcommand{\se}[1]{{\scriptsize$\pm#1$}}

\newcounter{insight}

\usepackage{todonotes}

\usepackage{titlesec}
\titlespacing*{\paragraph}{0pt}{0.25ex}{2ex}

\definecolor{chaptercolor}{HTML}{1A254B}
\definecolor{darkblue}{HTML}{1A254B}
\definecolor{linkcolor}{HTML}{2B50AA}
\definecolor{citecolor}{HTML}{2B50AA}
\definecolor{lightlinkcolor}{HTML}{9A8F97}
\definecolor{darklinkcolor}{HTML}{1A254B}
\definecolor{light}{HTML}{F8F8F8}
\definecolor{lightblue}{HTML}{A7BED3}
\definecolor{red}{HTML}{F2545B}
\definecolor{blue}{HTML}{2b50aa}

\theoremstyle{plain}

\newcommand{\figref}[2]{Figure~\hyperref[#1]{\ref{#1} (#2)}}

\usepackage{import}
\usepackage{xifthen}
\usepackage{pdfpages}
\usepackage{transparent}

\NewDocumentCommand{\incfig}{mo}{
  \begin{center}
    \IfValueT{#2}{\def\svgwidth{#2}}{\def\svgwidth{\columnwidth}}
    \import{./figures/}{#1.pdf_tex}
  \end{center}
}

\usepackage{pgf}
\usepackage{adjustbox}

\NewDocumentCommand{\incplt}{O{\columnwidth}m}{%
  \begin{center}
    \adjustbox{center}{\adjustbox{width=#1+10pt}{\includegraphics[width=#1]{./plots/output/#2.pdf}}}
  \end{center}
}

\NewDocumentCommand{\norm}{sm}{\IfBooleanTF{#1}{\|#2\|}{\left\| #2 \right\|}}
\NewDocumentCommand{\normF}{sm}{\IfBooleanTF{#1}{\|#2\|_{\mathrm{F}}}{\left\| #2 \right\|_{\mathrm{F}}}}
\NewDocumentCommand{\dTV}{sm}{d_{\mathrm{TV}}\IfBooleanTF{#1}{(#2)}{\left( #2 \right)}}

\DeclarePairedDelimiter\parentheses{(}{)}
\DeclarePairedDelimiter\brackets{[}{]}
\DeclarePairedDelimiter\braces{\{}{\}}

\NewDocumentCommand{\irred}{som}{\ensuremath{\sigma_{\hspace{-1pt}\infty}\IfBooleanTF{#1}{^2}{}(#3\IfValueTF{#2}{;#2}{})}}

\NewDocumentCommand{\fnPr}{}{\mathbb{P}}
\RenewDocumentCommand{\Pr}{om}{\fnPr\IfValueT{#1}{_{#1}}\parentheses*{#2}}
\RenewDocumentCommand{\H}{mo}{\mathrm{H}\IfValueTF{#2}{\!\left[#1\ \middle|\ #2\right]}{\brackets*{#1}}}
\NewDocumentCommand{\Hsm}{mo}{\mathrm{H}\IfValueTF{#2}{[#1 \mid #2]}{\brackets{#1}}}
\NewDocumentCommand{\I}{mmo}{\mathrm{I}\IfValueTF{#3}{\!\left(#1;#2\ \middle|\ #3\right)}{\parentheses*{#1; #2}}}
\NewDocumentCommand{\Ism}{mmo}{\mathrm{I}\IfValueTF{#3}{(#1;#2 \mid #3)}{\parentheses{#1; #2}}}

\NewDocumentCommand{\E}{somo}{\ensuremath{\mathbb{E}\IfValueT{#2}{_{#2}}{} \IfBooleanTF{#1}{#3}{\IfValueTF{#4}{\!\left[#3\ \middle|\ #4\right]}{\brackets*{#3}}}}}
\NewDocumentCommand{\Esm}{somo}{\ensuremath{\mathbb{E}\IfValueT{#2}{_{#2}}{} \IfBooleanTF{#1}{#3}{\IfValueTF{#4}{\!\left[#3\ \middle|\ #4\right]}{\brackets{#3}}}}}
\NewDocumentCommand{\Var}{somo}{\mathrm{Var}\IfValueT{#2}{_{#2}}{} \IfBooleanTF{#1}{#3}{\IfValueTF{#4}{\!\left(#3\ \middle|\ #4\right)}{\parentheses*{#3}}}}
\NewDocumentCommand{\Varsm}{somo}{\mathrm{Var}\IfValueT{#2}{_{#2}}{} \IfBooleanTF{#1}{#3}{\IfValueTF{#4}{\left(#3\ \middle|\ #4\right)}{\parentheses{#3}}}}
\NewDocumentCommand{\Cov}{som}{\mathrm{Cov}\IfValueT{#2}{_{#2}}{} \IfBooleanTF{#1}{#3}{\brackets*{#3}}}
\NewDocumentCommand{\Cor}{som}{\mathrm{Cor}\IfValueT{#2}{_{#2}}{} \IfBooleanTF{#1}{#3}{\brackets*{#3}}}

\NewDocumentCommand{\grad}{e_}{\boldsymbol{\nabla}\IfValueT{#1}{_{\!\!#1}\,}}

\NewDocumentCommand{\diag}{som}{\mathrm{diag}\IfValueT{#2}{_{#2}}{} \IfBooleanTF{#1}{\braces{#3}}{\braces*{#3}}}

\NewDocumentCommand{\N}{somm}{\mathcal{N}\IfBooleanTF{#1}{\left(}{(}\IfValueT{#2}{#2;}{} #3, #4\IfBooleanTF{#1}{\right)}{)}}
\NewDocumentCommand{\GP}{omm}{\mathcal{GP}(\IfValueT{#1}{#1;}{} #2, #3)}

\begin{document}

\ifcolmsubmission
\linenumbers
\fi

\maketitle

\begin{abstract}
    Recent work has shown that language models can self-improve by maximizing their own confidence in their predictions, without relying on external verifiers or reward signals.
    In this work, we study the test-time scaling of language models for mathematical reasoning tasks, where the model's own confidence is used to select the most promising attempts.
    Surprisingly, we find that we can achieve significant performance gains by continuing only the most promising attempt, selected by the model's \emph{prefix-confidence}.
    We systematically evaluate prefix-confidence scaling on five mathematical reasoning datasets: the school-level GSM8K and MATH500, and the competition-level AMC23, AIME24, and AIME25.
    We find that prefix-confidence scaling with prefixes of only $32$ tokens achieves a better accuracy-compute trade-off than majority voting. Moreover, prefix-confidence scaling appears less susceptible than Bo$N$ to length biases.
    Finally, we also evaluate test-time training with prefix-confidence and find that, while outperforming the base model, it does not improve over prefix-confidence scaling.\looseness=-1
\end{abstract}

\begin{figure}[H]
    \vspace{-2ex}
    \centering
    \begin{minipage}[c]{0.49\textwidth}
        \centering
        \includegraphics[width=\linewidth]{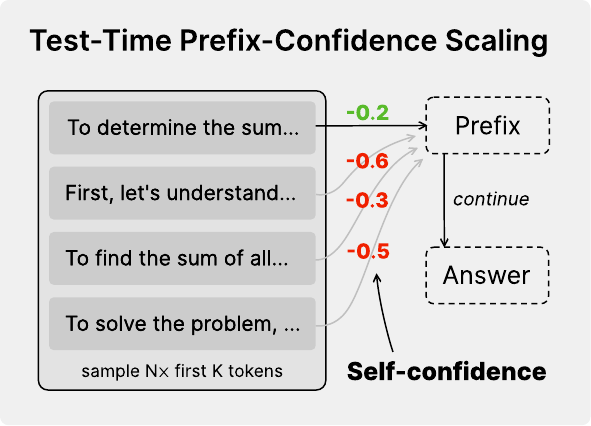}
    \end{minipage}
    \hfill
    \begin{minipage}[c]{0.49\textwidth}
        \centering
        \includegraphics[width=\linewidth]{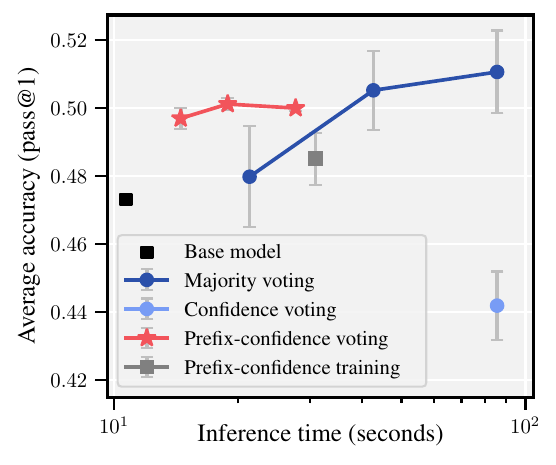}
        \vspace{-4ex}
    \end{minipage}
    \caption{\textbf{Left:}~We study test-time scaling of language models in mathematical reasoning \emph{without a verifier}, by instead maximizing the model's own confidence in its predictions. We find that models can self-improve by continuing the most confident attempt.
    \textbf{Right:}~We evaluate prefix-confidence scaling on GSM8K, MATH500, AMC23, AIME24, and AIME25 with Qwen2.5-Math-1.5B-Instruct as base model.
    \textbf{Compared to majority voting and best-of-$N$ (``confidence voting'' on full attempts), prefix-confidence scaling achieves a better accuracy-compute trade-off.} We further find that leveraging prefix-confidence for test-time inference (``voting'') performs better than test-time training (``training'') at matched latency. Error bars show the standard error across 10 seeds.\looseness=-1}
    \label{fig:main}
    \vspace{-1ex}
\end{figure}

\section{Introduction}
\vspace{-1ex}

Self-improving language models have been the center of attention of many recent works~\citep[e.g.,][]{huang2023large,huang2025self,lee2025self,zweiger2025self,zhao2025absolute}.
The idea of a language model that is able to improve its own performance \emph{without relying on any external verifier or reward signal} is appealing for many reasons.
In particular, such self-improvement may be applicable to a wide range of tasks, including those where no explicit reward signal is available.
In this work, we focus on challenging mathematical reasoning tasks, where the model does not have access to an external environment that checks the correctness of its attempts.
Instead of relying on feedback from an environment, previous approaches to self-improvement in language models commonly elicit a model's notion of confidence in its own predictions to determine the most promising attempts.
While many measures of confidence, such as majority voting~\citep[e.g.,][]{snell2025scaling}, (negative) entropy~\citep[e.g.,][]{agarwal2025unreasonable} or ``self-certainty''~\citep{kang2025scalable} have been proposed, they are all based entirely on the base model.
As part of this work, we survey common measures of confidence.\looseness=-1

Prior work studying maximizing the model's own confidence has done so either at test-time via majority voting or best-of-$N$~(Bo$N$) sampling~\citep[e.g.,][]{huang2025self}, or at train-time via supervised fine-tuning~\citep{huang2023large,huang2025self,agarwal2025unreasonable,li2025confidence} or reinforcement learning~\citep{agarwal2025unreasonable,prabhudesai2025maximizing,zuo2025ttrl,huang2025self,zhao2025learning,xu2025direct}.
In this work, we study the test-time scaling of maximizing self-confidence.
In particular, we ask whether self-confidence can be optimized \emph{more efficiently} than by majority voting or Bo$N$ sampling, which require generating multiple full attempts.
\textbf{Instead of generating multiple full attempts, we find that performance improves significantly when continuing only the most promising prefix, selected according to the model's self-confidence.}
We refer to this method as \emph{test-time prefix-confidence scaling}.
We evaluate two confidence measures: ``self-confidence''~\citep{huang2025self} and ``self-certainty''~\citep{kang2025scalable}, and find that self-confidence tends to outperform self-certainty.
Moreover, we evaluate test-time inference and test-time training as methods for optimizing prefix-confidence, and find that test-time inference tends to outperform test-time training.\looseness=-1

\vspace{-1ex}
\section{Related Work}\label{sec:related_work}
\vspace{-1ex}

\paragraph{Test-time inference without verifiers.}

A practical and verifier-free strategy for scaling test-time computation is to aggregate attempts directly using simple heuristics, without relying on explicit learned reward models.
A common approach is \emph{majority voting}~(also called \textbf{self-consistency}), where the model generates several independent completions and the final output is selected by consensus among them~\citep{snell2025scaling}.
Majority voting, however, has the disadvantage of requiring explicit answers, meaning that it cannot generally be used in open-ended tasks beyond mathematical reasoning.
Alternatively, \emph{best-of-$N$ sampling}, which selects the response with highest self-confidence (based on some confidence measure), has been shown to improve performance on certain tasks~\citep{huang2025self,kang2025scalable}.
Beyond naive sampling, structured search procedures such as beam search further explore multiple promising attempts.
Nevertheless, prior test-time inference methods are centered around generating multiple complete attempts.
In contrast, in this work, we evaluate selecting the most promising prefix, and only fully generating a single attempt.\looseness=-1

\paragraph{Test-time training.}

An alternative approach to test-time scaling is to continue training the language model at test-time, known as test-time training~\citep[TTT,][]{sun2020test}.
Approaches vary primarily in the loss they utilize for TTT.
Some approaches, also known as dynamic evaluation, train the model with a self-supervised imitation loss on the token stream~\citep{krause2018dynamic,krause2019dynamic,sun2024learning,dalal2025one,hu2025test}.
Other methods train on a self-supervised imitation loss on ``expert data'', such as few-shot examples~\citep{akyurek2024surprising} or examples from a training corpus similar to the prompt~\citep{hardt2024test,hubotter2024efficiently,bertolissi2025local}.
An orthogonal line of work studies TTT without imitating any expert data, and instead maximizing the model's confidence in its predictions. Prominent examples are TENT~\citep{wang2021tent} and MEMO~\citep{zhang2022memo}, which minimize the entropy of the predictive distribution of a classifier on a batch of test points.\looseness=-1

\paragraph{Maximizing confidence.}

Several works have proposed different self-confidence measures, and applied them for test-time inference, supervised fine-tuning~(SFT) or reinforcement learning~(RL). In the following, we summarize common confidence measures.
\begin{itemize}[leftmargin=6ex]
    \item \textbf{Self-consistency} (majority voting) is often used at inference-time~\citep{snell2025scaling}, but has also been used with SFT~\citep{huang2023large,lee2025self} and RL~\citep{zuo2025ttrl,yuan2025reinforce}.

    \item \textbf{Self-confidence} measures the likelihood of the attempt~$y \sim \pi(\cdot \mid x)$ to question~$x$: \begin{align}
        \log \pi(y \mid x) = \textstyle\sum_{i=1}^n \log \pi(y_i \mid x, y_{<i}), \label{eq:self_confidence}
    \end{align} where $n$ denotes the length of the attempt.
    This has been used for test-time inference, SFT~\citep{huang2025self}, and RL~\citep{huang2025self,zhao2025learning,xu2025direct}. Self-confidence is typically measured without scaling by $1/n$, but is susceptible to length biases, as longer sequences tend to have lower self-confidence.

    \item \textbf{Self-certainty} (measured on the token-level) maximizes the KL-divergence of the model's predictions to a uniform distribution (i.e., maximum uncertainty): \begin{align}
    \tfrac{1}{n} \textstyle\sum_{i=1}^n \mathrm{KL}(\mathrm{Unif} \| \pi(y_i \mid x, y_{<i})) = -\tfrac{1}{n} \textstyle\sum_{i=1}^n \textstyle\sum_{j=1}^V \log \pi(y_i = j \mid x, y_{<i}) + \mathrm{const}, \label{eq:self_certainty}
    \end{align} where $V$ denotes the vocabulary size.
    This been used for test-time inference~\citep{kang2025scalable} and RL~\citep{zhao2025learning}, but may also be susceptible to length biases.

    \item \textbf{Negative entropy} additionally weights each term in \cref{eq:self_certainty} by the probability of the generated token~(cf.~\cref{eq:entropy_confidence} in \cref{sec:addtl_confidence_measures}).
    Entropy has been used for SFT~\citep{agarwal2025unreasonable,li2025confidence} and RL~\citep{agarwal2025unreasonable,prabhudesai2025maximizing,zhang2025right}. We do not evaluate entropy in this work, since prior work has shown it to be outperformed by self-certainty~\citep{kang2025scalable}.
\end{itemize}
Finally, our work is closely related to recent work on \emph{prefix-tuning}~\citep{ji2025first}, which studies the effectiveness of training the model on its own prediction prefixes to stabilize and align early reasoning across solutions.
As opposed to \cite{ji2025first}, we do not train the model on its own prefixes (at train-time), but rather evaluate the effectiveness of leveraging prefixes at test-time for test-time scaling.\looseness=-1

\vspace{-1ex}
\section{Method and Results}
\vspace{-1ex}

We evaluate whether we can effectively maximize the model's confidence based on just prefixes, as opposed to using full attempts as in majority voting or best-of-$N$ sampling.
To this end, we explore two approaches: test-time inference and test-time training. \begin{itemize}[leftmargin=6ex]
    \item \textbf{Prefix-confidence voting} (test-time inference): We generate $N$ prefixes of length $K$ with the base model, and select the most confident prefix according to the model's self-confidence~(cf.~\cref{eq:self_confidence}) or self-certainty~(cf.~\cref{eq:self_certainty}).

    \item \textbf{Prefix-confidence training} (test-time training): We train the model on the $N$ prefixes of length $K$ generated by the base model. As loss, we evaluate the standard negative log-likelihood (NLL) loss~(cf.~\cref{eq:nll_loss} detailed in \cref{sec:ttt_losses}) similarly to \cite{agarwal2025unreasonable}, and a loss that minimizes the entropy of the generated sequences~(cf.~\cref{eq:entropy_loss} detailed in \cref{sec:ttt_losses}).
\end{itemize}

\paragraph{Base model and datasets.}

We use the Qwen-2.5-Math-1.5B-Instruct model~\citep{yang2024qwen2} as base model, which is state-of-the-art for mathematical reasoning tasks at its scale.
We evaluate on five mathematical reasoning datasets: the school-level GSM8K~\citep{cobbe2021training} and MATH500~\citep{hendrycks2021measuring,lightman2023let}, and the competition-level AMC23, AIME24, and AIME25.
We use the official test splits of the datasets, except for GSM8K where we only evaluate on the first $500$ examples of the test set.
For each experiment, we report standard errors across $10$ random seeds. We measure inference time on an NVIDIA RTX 4090 GPU.\footnote{We report the inference time corresponding to the average query/answer per dataset.}\looseness=-1

\begin{figure}
    \vspace{-3ex}
    \centering
    \incplt[\linewidth]{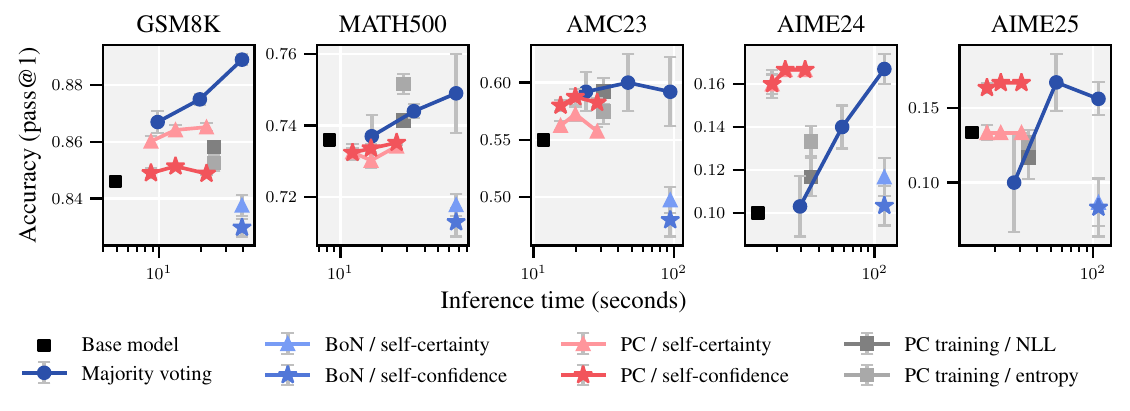}
    \vspace{-3ex}
    \caption{We evaluate prefix-confidence~(PC) scaling with prefix length $32$ on five mathematical reasoning datasets.
    We vary the number of samples for PC voting across $8, 16, 32$. As baselines, we evaluate majority voting on $2, 4, 8$ samples and Bo$N$ on $8$ samples. We further evaluate PC training with $32$ samples.
    We find that both PC voting and PC training tend to outperform the base model significantly, with PC voting typically outperforming PC training.
    Furthermore, except for GSM8K, self-confidence outperforms self-certainty.
    Notably, we find that confidence maximization with Bo$N$ on full attempts almost always performs \emph{worse} than the base model, which we attribute to length biases.
        \looseness=-1}
    \label{fig:per_task}
\end{figure}

\begin{table}[t]
  \centering
  \small
  \begin{tabular}{@{} l >{\scriptsize}l *{5}{c} c c @{}}
    \toprule
     & & GSM8K  & MATH500 & AMC23  & AIME24 & AIME25 & \textbf{avg}       & time (s) \\
    \midrule
    Base
      &        & 84.6   & 73.6    & 55.0   & 10.0   & 13.3   & 47.3  & 10.68 \\
    \midrule

    \multirow{2}{*}{Bo$N$@8}
      & self-confidence
                 & 83.0\se{0.3} & 71.3\se{0.4} & 48.0\se{1.4} & 10.3\se{0.9} &  8.3\se{1.9}
                 & 44.2\se{0.5} & \multirow{2}{*}{85.43} \\
      & self-certainty
                 & 83.8\se{0.4} & 71.8\se{0.3} & 49.8\se{1.1} & \underline{11.7}\se{0.9} &  8.7\se{1.6}
                 & 45.2\se{0.4} & \\

    \addlinespace
    \multirow{2}{*}{Bo$N$@16}
      & self-confidence
                 & 82.0\se{0.2} & 69.7\se{0.2} & 44.8\se{1.2} &  9.0\se{1.3} &  5.3\se{0.9}
                 & 42.2\se{0.4} & \multirow{2}{*}{170.86} \\
      & self-certainty
                 & 83.0\se{0.2} & 70.2\se{0.4} & 47.5\se{1.4} &  9.7\se{1.0} &  6.3\se{1.8}
                 & 43.3\se{0.5} & \\

    \addlinespace
    Maj@2
      & self-consistency
                 & \underline{86.7}\se{0.4} & 73.7\se{0.6} & \underline{\textbf{59.2}}\se{1.7} & 10.3\se{1.4} & 10.0\se{3.3}
                 & 48.0\se{0.8} & 21.36 \\
    Maj@4
      & self-consistency
                 & \underline{87.5}\se{0.2} & \underline{\textbf{74.4}}\se{0.2} & \underline{\textbf{60.0}}\se{2.5} & \underline{14.0}\se{1.0} & \underline{\textbf{16.7}}\se{1.9}
                 & \underline{\textbf{50.5}}\se{0.7} & 42.72 \\
    Maj@8
      & self-consistency
                 & \underline{\textbf{88.9}}\se{0.2} & \underline{\textbf{74.9}}\se{1.1} & \underline{\textbf{59.2}}\se{3.0} & \underline{\textbf{16.7}}\se{0.7} & \underline{\textbf{15.6}}\se{1.1}
                 & \underline{\textbf{51.1}}\se{0.7} & 85.43 \\

    \midrule\rowcolor{lightgray}
    \cellcolor{white}{}%
    \multirow{2}{*}{PC@8}
      & self-confidence
                 & \underline{84.9}\se{0.2} & 73.2\se{0.1} & \underline{58.0}\se{0.5} & \underline{16.0}\se{0.4} & \underline{16.3}\se{0.3}
                 & \underline{49.7}\se{0.1} & \cellcolor{white}{}\multirow{2}{*}{14.52} \\
      & self-certainty
                 & \underline{\textbf{86.0}}\se{0.2} & 73.3\se{0.2} & \underline{56.3}\se{0.4} & \underline{\textbf{16.0}}\se{0.7} & 13.3\se{0.5}
                 & \underline{49.0}\se{0.2} & \\

    \addlinespace\rowcolor{lightgray}
    \cellcolor{white}{}%
    \multirow{2}{*}{PC@16}
      & self-confidence
                 & \underline{85.1}\se{0.1} & 73.4\se{0.1} & \underline{\textbf{58.8}}\se{0.7} & \underline{\textbf{16.7}}\se{0.0} & \underline{\textbf{16.7}}\se{0.0}
                 & \underline{\textbf{50.1}}\se{0.1} & \cellcolor{white}{}\multirow{2}{*}{18.89} \\
      & self-certainty
                 & \underline{\textbf{86.4}}\se{0.2} & 73.0\se{0.2} & \underline{57.3}\se{0.6} & \underline{\textbf{16.7}}\se{0.0} & 13.3\se{0.0}
                 & \underline{49.3}\se{0.1} & \\

    \addlinespace
    \multirow{2}{*}{PC training}
      & NLL~\eqref{eq:nll_loss}
                 & \underline{85.8}\se{0.2} & \underline{74.1}\se{0.2} & \underline{\textbf{59.3}}\se{1.1} & \underline{11.7}\se{0.9} & 11.7\se{1.4}
                 & \underline{48.5}\se{0.4} & \multirow{2}{*}{30.87} \\
      & entropy~\eqref{eq:entropy_loss}
                 & \underline{85.3}\se{0.3} & \underline{\textbf{75.2}}\se{0.3} & \underline{57.5}\se{1.1} & \underline{13.3}\se{0.7} & 12.7\se{0.8}
                 & \underline{48.8}\se{0.3} & \\
    \bottomrule
  \end{tabular}
  \caption{Accuracy across mathematical reasoning datasets. \textbf{Bold} numbers denote the best performance across either non-prefix- or prefix-based methods. \underline{Underlined} numbers denote if a method outperforms the base model. All prefix-based methods use prefix length~$32$.\looseness=-1}
  \label{tab:main}
  \vspace{-2ex}
\end{table}

\paragraph{Finding: Maximizing prefix-confidence efficiently improves mathematical reasoning.}

We detail our results in \cref{fig:per_task} and \cref{tab:main}.
We find that prefix-confidence voting tends to perform optimally on the accuracy-compute frontier, in particularly on harder and newer tasks.
Surprisingly, we find that Bo$N$ on full attempts does not improve upon the base model.
We hypothesize that this is due to inevitable length biases when measuring confidence across sequences of different lengths.
Prefix-confidence scaling has the advantage that all compared prefixes are of the same length, avoiding such biases.\looseness=-1

We ablate the choice of prefix length $K$ in \cref{fig:prefix_length} in the appendix.
Similarly to \cite{ji2025first}, we find that prefixes of length up to 32 are sufficient for prefix-confidence scaling.
We further ablate prefix-confidence training in \cref{sec:ablataions_ttt}.\looseness=-1

\vspace{-1ex}
\section{Conclusion}
\vspace{-1ex}

We study test-time scaling based on the model's confidence in its own prefixes.
Unlike methods selecting between multiple full attempts, such as Bo$N$, prefix-confidence scaling is substantially more efficient and does not suffer from length biases.
In contrast to majority voting, prefix-confidence does not require explicit answers, hence, evaluation outside mathematical reasoning is an exciting direction for future work.
Other interesting directions are to select the prefix length dynamically based on the question difficulty or to use self-consistency as a confidence measure on prefixes, where prefixes are grouped into latent semantic categories for voting, akin to \cite{zhang2025right}.\looseness=-1

\section*{Acknowledgments}

This project was supported in part by the European Research Council (ERC) under the European Union's Horizon 2020 research and Innovation Program Grant agreement no. 815943, and the Swiss National Science Foundation under NCCR Automation, grant agreement 51NF40 180545. Ido Hakimi was supported by an ETH AI Center Postdoctoral fellowship.\looseness=-1

\bibliography{colm2025_conference}
\bibliographystyle{colm2025_conference}

\clearpage\appendix

\begin{figure}[t]
    \centering
    \incplt[0.5\linewidth]{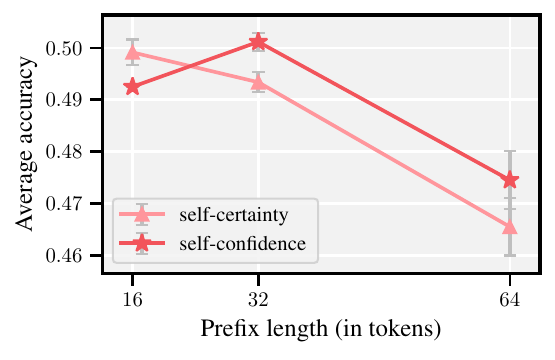}
    \caption{Comparison of prefix-confidence voting across prefix lengths, averaged over datasets.
        \looseness=-1}
    \label{fig:prefix_length}
\end{figure}

\section{Additional Confidence Measures}\label{sec:addtl_confidence_measures}

\begin{itemize}[leftmargin=6ex]
    \item \textbf{Negative entropy} (token-level) \begin{align}
        -\tfrac{1}{n} \textstyle\sum_{i=1}^n \mathrm{H}(\pi(y_i \mid x, y_{<i})) = -\tfrac{1}{n} \textstyle\sum_{i=1}^n \textstyle\sum_{j=1}^V \pi(y_i = j \mid x, y_{<i}) \log \pi(y_i = j \mid x, y_{<i}). \label{eq:entropy_confidence}
    \end{align}
\end{itemize}

\section{Test-time Training Losses}\label{sec:ttt_losses}

For $K$ prefixes $\{y^k \sim \pi_\theta(\cdot \mid x)\}_{k=1}^K$, we evaluate the following losses for prefix-confidence training: \begin{itemize}[leftmargin=6ex]
    \item \textbf{Negative log-likelihood loss}: \begin{align}
        L(\theta \mid x) := - \sum_{k=1}^K \log \pi_\theta(y^k \mid x) \label{eq:nll_loss}
    \end{align}

    \item \textbf{Entropy loss} (on the trajectory-level): \begin{align}
        L(\theta \mid x) :\!\!&= - \sum_{k=1}^K \pi_\theta(y^k \mid x) \log \pi_\theta(y^k \mid x) \label{eq:entropy_loss} \\
        &\approx K \cdot \mathbb{E}_{y \sim \pi_\theta(\cdot \mid x)}[- \log \pi_\theta(y \mid x)] = K \cdot \mathrm{H}(\pi_\theta(y \mid x))
    \end{align}
\end{itemize}

\section{Qualitative Example}

\begin{table}[H]
   \centering
   \begin{tabular}{>{\raggedright\arraybackslash}p{\textwidth}}
   \toprule
   \textbf{Question:} Find the sum of all integer bases $b>9$ for which $17_b$ is a divisor of $97_b.$ \\
   \midrule
       To determine the sum of all integer bases \( b > 9 \) for which \( 17_b \) is a divisor of \dots
       \\
       First, let's understand the given problem. We have the numbers \(17_b\) and \(97_b\) in base \dots
       \\
       To find the sum of all integer bases \( b > 9 \) for which \( 17_b \) is a divisor of \dots
       \\
       To solve the problem, we need to convert the numbers \(17_b\) and \(97_b\) from base \(b\) \dots
       \\
   \bottomrule
   \end{tabular}
   \caption{Example of the first 32 tokens of responses to a question from AIME 2025}
   \label{table:example_generations}
\end{table}

\section{Ablations of Prefix-Confidence Training}\label{sec:ablataions_ttt}

In this section, we ablate the choice of hyperparameters for prefix-confidence training.
We denote by $K$ the prefix length (number of tokens), by $N$ the number of sampled prefixes, and by $E$ the number of epochs.

\subsection{Number of Gradient Steps}
\label{sec:gradient_steps}

We do not use batching for the fine-tuning of our model, and thus always take $N \cdot E$ gradient steps for each test point.

\begin{figure}[t]
    \centering
    \begin{subfigure}[b]{0.48\textwidth}
        \centering
        \includegraphics[width=\textwidth]{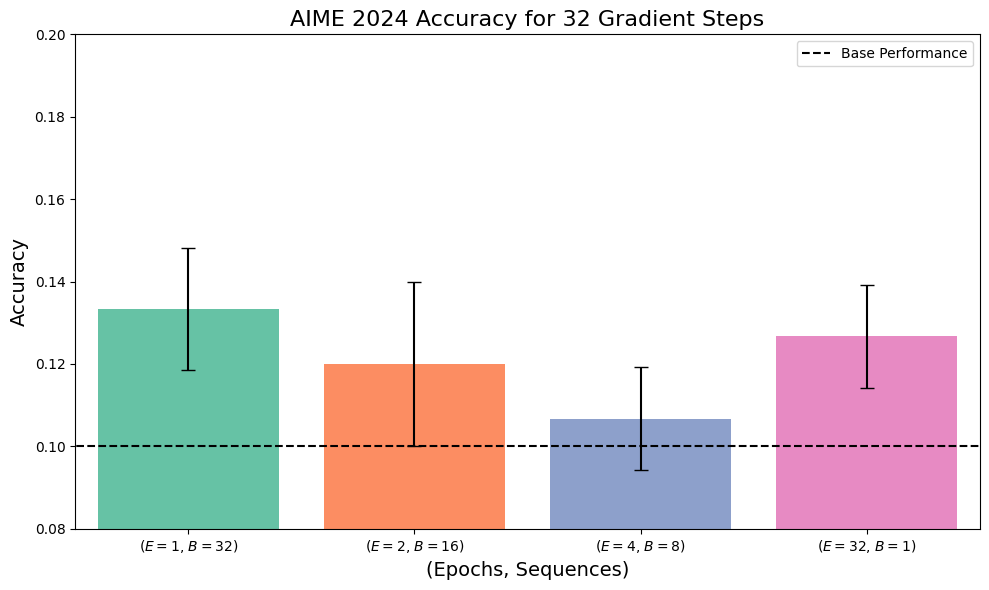}
        \caption{AIME 2024}
        \label{fig:bve_aime}
    \end{subfigure}
    \begin{subfigure}[b]{0.48\textwidth}
        \centering
        \includegraphics[width=\textwidth]{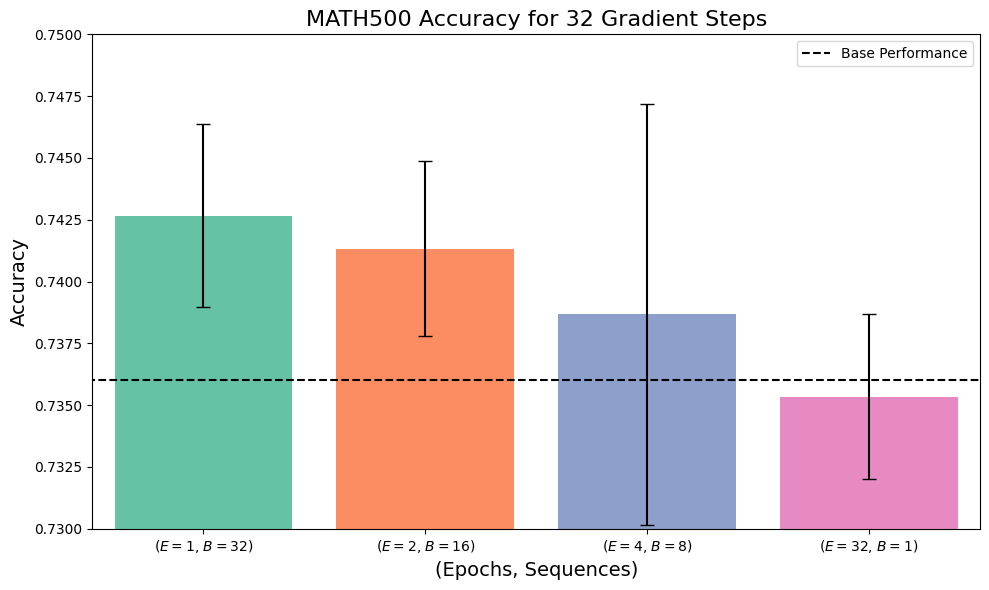}
        \caption{MATH500}
        \label{fig:bve_math}
    \end{subfigure}
    \caption{Comparison of performance on AIME24 and MATH500 for varying $E$. We show (accuracy - best accuracy), so the accuracy improvement. Here, $B = N$.}
    \label{fig:bve}
\end{figure}

In \cref{fig:bve}, we show the effect of different allocations of the gradient steps on the AIME24 and MATH500 dataset.
We examine the setting of $N \cdot E = 32$, with different choices of $N$ and $E$.
We can see that the trend between the two datasets differs, but in both cases using $32$ samples with just a single epoch performs best.
Based on this, we fix $E=1$ for the rest of the experiments.

\subsection{Number of Samples}

Next, we investigate the number of samples $N$ we use for prefix-confidence training.
Similarly to \cite{ji2025first}, we fix the prefix length to $K=32$ tokens.

\begin{figure}[t]
    \centering
    \begin{subfigure}[b]{0.48\textwidth}
        \centering
        \includegraphics[width=\textwidth]{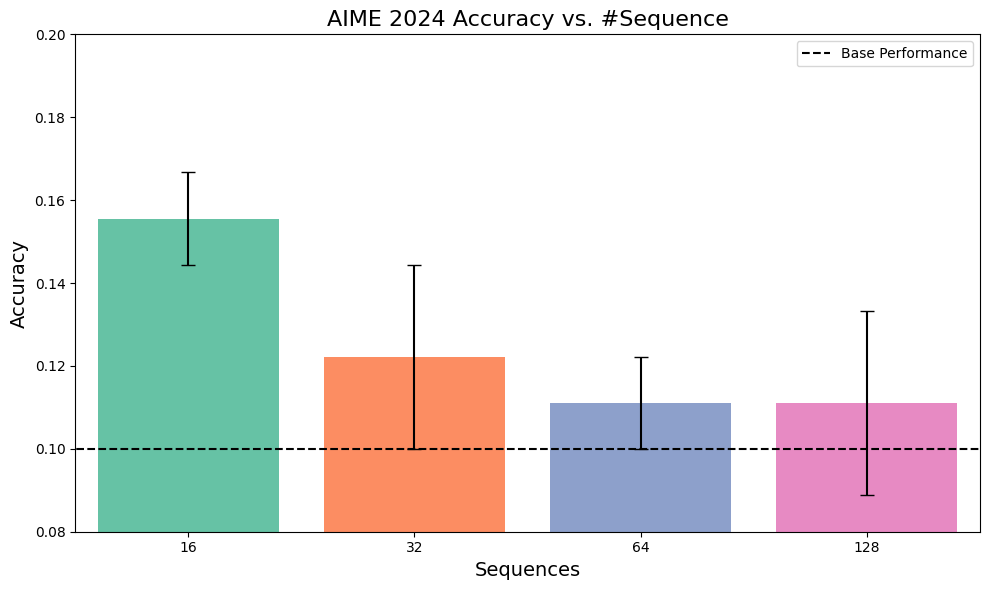}
        \caption{AIME 2024}
        \label{fig:seqs_aime}
    \end{subfigure}
    \begin{subfigure}[b]{0.48\textwidth}
        \centering
        \includegraphics[width=\textwidth]{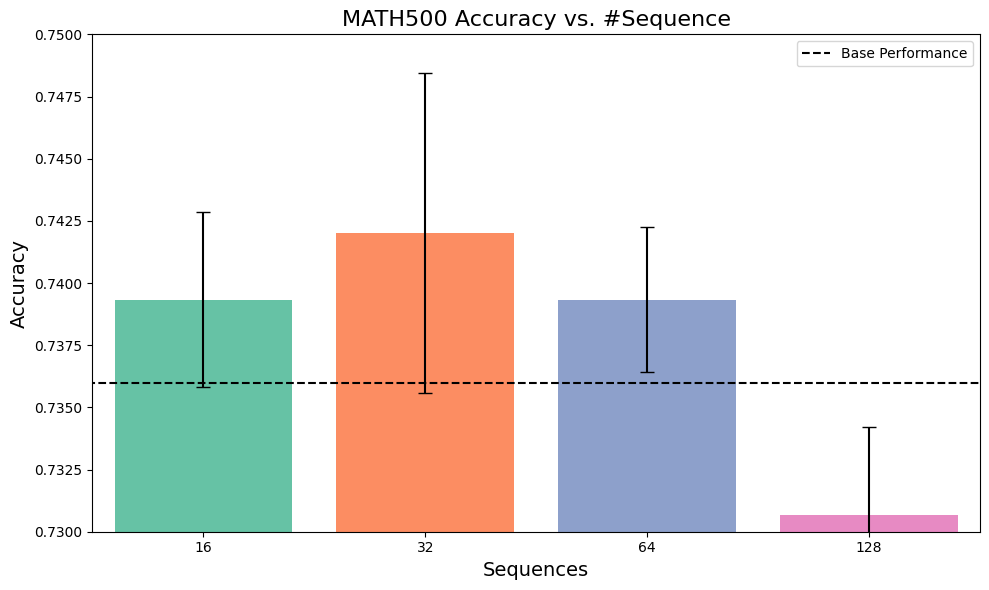}
        \caption{MATH500}
        \label{fig:seqs_math}
    \end{subfigure}
    \caption{Comparison of performance on AIME24 and MATH500 for varying $N$. We show (accuracy - best accuracy), so the accuracy improvement.}
    \label{fig:budget}
\end{figure}

In \cref{fig:budget}, we show the effect of different values for $N$.
The two datasets show a slightly different trend.
Since AIME24 is much smaller than MATH500, we choose $N=32$ for the rest of the ablations.
This results in $32 \cdot 32 = 1024$ gradient steps for each test point.

\subsection{Optimizer}
\label{sec:optimizer}

\begin{figure}[t]
    \centering
    \begin{subfigure}[b]{0.48\textwidth}
        \centering
        \includegraphics[width=\textwidth]{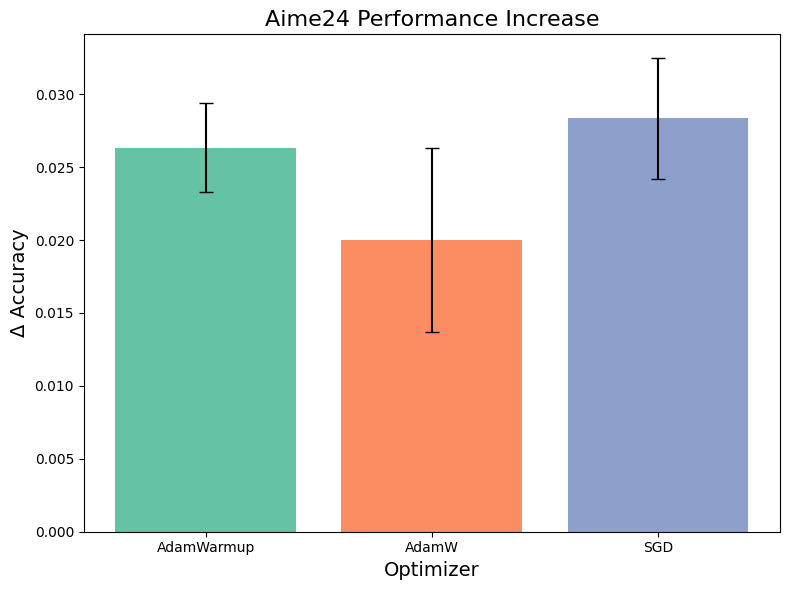}
        \caption{AIME 2024}
        \label{fig:adam}
    \end{subfigure}
    \begin{subfigure}[b]{0.48\textwidth}
        \centering
        \includegraphics[width=\textwidth]{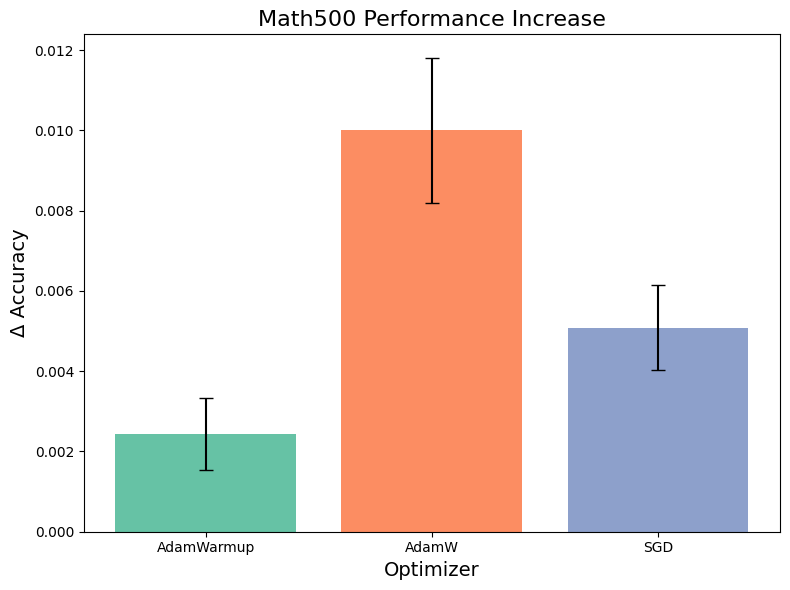}
        \caption{MATH500}
        \label{fig:adamwarmup}
    \end{subfigure}
    \caption{Comparison of different optimizers on AIME24 and MATH500. We show (accuracy - best accuracy), so the accuracy improvement.}
    \label{fig:optimizers}
\end{figure}

We consider three optimizers: SGD, AdamW~\citep{kingma2015adam,loshchilov2019decoupled}, and a version of AdamW (which we call AdamWarmup) where we warmup the moments based on a single generated prefix.
For each optimizer, we first ablated over the learning rate.
Only the best learning rate is shown in \cref{fig:optimizers}.
For AdamW and AdamWarmup, we used a learning rate of $5e-6$ and for SGD we used a learning rate of $2e-3$.
We find that the best optimizer is not always the same for each dataset.
Since the SGD optimizer uses less memory and compute, we continued using SGD for the rest of the experiments.

\subsection{Incorporating Full Attempts}

For a fraction $r = 0.1$ of the training data, \cite{ji2025first} generate the full reasoning trace to prevent catastrophic forgetting.
This fraction $r$ is called the tuning ratio.

\begin{figure}[ht]
    \centering
    \includegraphics[width=0.7\textwidth]{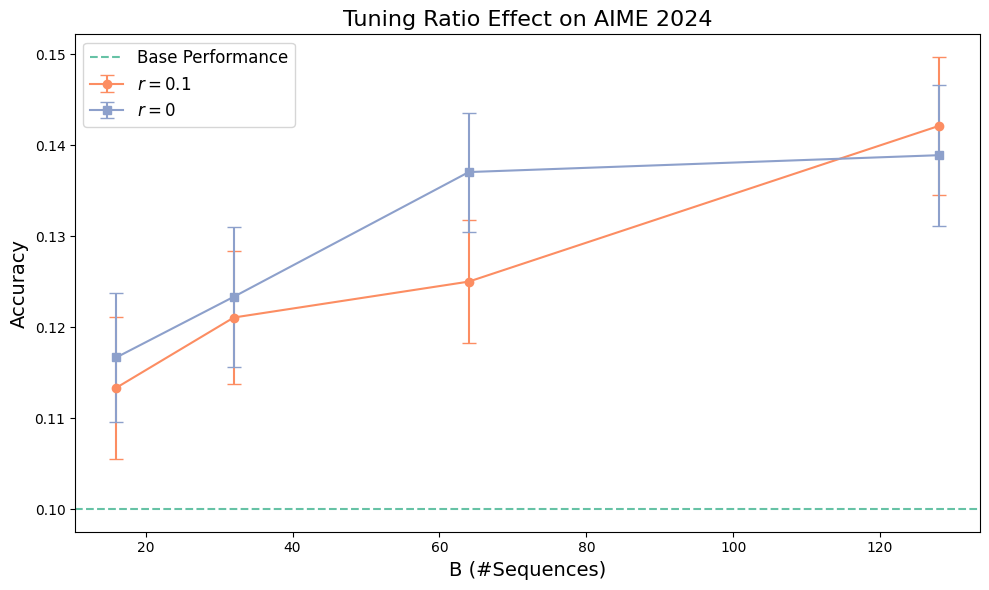}
    \caption{Showing the impact of the tuning ratio $r$ on AIME24. We compare runs with different numbers of sampled sequences for fine-tuning, to show the impact of the tuning ratio with increasing budget.
    The standard error is shown across 20 seeds.}
    \label{fig:tuning_ratio}
\end{figure}

We ablate the effect of the tuning ratio in \cref{fig:tuning_ratio}.
We find that the tuning ratio does not have a significant impact on the performance.
This is likely because we only ever perform few gradient steps on top of the base model, and thus catastrophic forgetting does not occur.
As incorporating full attempts is computationally expensive, we set $r=0$ for the rest of the experiments.

Additionally, we find on AIME24 that increasing the number of sequences $N$ beyond $32$ leads to further performance improvements. We leave a more thorough investigation of $N$'s effect depending on the dataset to future work.

\end{document}